\documentclass[conference]{IEEEtran}
\IEEEoverridecommandlockouts

\usepackage{amsmath,amssymb,amsfonts}
\usepackage{graphicx}
\usepackage{textcomp}
\usepackage{xcolor}
\usepackage{graphicx}
\usepackage{amsthm}
\usepackage{alltt}
\usepackage{mathrsfs}
\usepackage{subfigure}
\usepackage{enumerate}
\usepackage[noadjust]{cite}
\usepackage{float}
\usepackage{hyperref}
\hypersetup{hidelinks}

\usepackage{algorithm}
\usepackage{algpseudocode}

\DeclareGraphicsExtensions{.eps}
\usepackage{soul}
\usepackage{array}
\def\BibTeX{{\rm B\kern-.05em{\sc i\kern-.025em b}\kern-.08em
    T\kern-.1667em\lower.7ex\hbox{E}\kern-.125emX}}

\DeclareMathOperator*{\argmax}{arg\,max}

\newcommand{\mb}[1]{\mathbf{#1}}
\newcommand{\mc}[1]{\mathcal{#1}}
\newcommand{\mbb}[1]{\mathbb{#1}}

\newlength{\figWidth}
\begin{document}

\title{Wireless Channel Identification via Conditional Diffusion Model\\
}

\author{
\IEEEauthorblockN{
Yuan Li\IEEEauthorrefmark{1},
Zhong Zheng\IEEEauthorrefmark{1},
Chang Liu\IEEEauthorrefmark{2}, and
Zesong Fei\IEEEauthorrefmark{1}}
\IEEEauthorblockA{\IEEEauthorrefmark{1}School of Information and Electronics, Beijing Institute of Technology, Beijing, China}
\IEEEauthorblockA{\IEEEauthorrefmark{2} School of Information Engineering, Guangdong University of Technology, Guangzhou, China}
\IEEEauthorblockA{Emails: 3220235334@bit.edu.cn, zhong.zheng@bit.edu.cn, liuchang@gdut.edu.cn, feizesong@bit.edu.cn}
}

\maketitle

\begin{abstract}
The identification of channel scenarios in wireless systems plays a crucial role in channel modeling, radio fingerprint positioning, and transceiver design. Traditional methods to classify channel scenarios are based on typical statistical characteristics of channels, such as K-factor, path loss, delay spread, etc. However, statistic-based channel identification methods cannot accurately differentiate implicit features induced by dynamic scatterers, thus performing very poorly in identifying similar channel scenarios.
In this paper, we propose a novel channel scenario identification method, formulating the identification task as a maximum a posteriori (MAP) estimation. Furthermore, the MAP estimation is reformulated by a maximum likelihood estimation (MLE), which is then approximated and solved by the conditional generative diffusion model.
Specifically, we leverage a transformer network to capture hidden channel features in multiple latent noise spaces within the reverse process of the conditional generative diffusion model. These detailed features, which directly affect likelihood functions in MLE, enable highly accurate scenario identification. Experimental results show that the proposed method outperforms traditional methods, including convolutional neural networks (CNNs), back-propagation neural networks (BPNNs), and random forest-based classifiers, improving the identification accuracy by more than 10\%.
\end{abstract}


\section{Introduction}
Vehicular networks are expected to play a pivotal role in 6G communications, supporting a wide range of applications in intelligent transportation, smart cities, and autonomous systems. However, vehicle communications occur in diverse scenarios where channel conditions vary significantly due to dynamic environmental obstacles and vehicle movement. These scenario variations directly impact channel modeling \cite{CM,pdesign1} and necessitate adaptive transceiver designs to accommodate dynamic channel characteristics.
For instance, in massive MIMO beamforming, the precoding strategy must be adjusted based on scenario identification. The directivity of beams should be optimized in line-of-sight (LoS) environments, while in non-line-of-sight (NLoS) environments it is appropriate to design scattering-enhanced beams. Real-time adaptation to changing channel scenarios is therefore essential for ensuring reliable communication in vehicular networks. As a result, scenario identification has become a critical task for optimizing wireless system performance and enabling robust connectivity.

In the realm of wireless communication, scenario identification is traditionally achieved via two primary methods: visual-based and channel information-based approaches.
Visual-based approaches rely on external sensors, such as cameras and LiDAR scanners, to capture imagery of the surrounding environment. While effective, they require specialized hardware, leading to high costs and limited compatibility with existing network infrastructure.

In contrast, channel information-based approaches exploit intrinsic information already available in communication systems. By analyzing statistical channel characteristics, such as the correlation of path clusters and time-varying angular features, numerous studies have focused on identifying the availability of LoS and NLoS conditions in a given propagation environment. For instance, support vector machines (SVMs) \cite{ANNLoS},  convolutional neural networks (CNNs) \cite{CNN}, and random forests \cite{AiboV2V, RF} have been utilized to differentiate LoS/NLoS conditions in indoor or vehicle-to-vehicle environments. Additionally, CNNs have been applied in \cite{IndoorUWBCNN} to identify LoS, weak LoS, and NLoS conditions in indoor communication.
Compared to simple binary identification between LoS and NLoS conditions, it is more complex to differentiate communication scenarios among indoor, urban, suburban, and others. In \cite{RFIndoorOutdoor}, random forests have been used to differentiate indoor and outdoor communication scenarios.
Moreover, in \cite{BPNN2021}, four typical scenarios, including urban, highways, tunnels, and vehicle obstructions, are identified using power delay spectrum and K-factors, via a back propagation neural network (BPNN). Similarly, based on path loss, K-factor, and delay spread, the authors of \cite{hybridSVMGMM} combined SVM and Gaussian mixture models to identify hybrid LoS/NLoS urban and suburban scenarios.

Despite their success, existing identification methods based on statistical channel characteristics rely heavily on the delicate selection of statistical channel characteristics. However, the differences in statistical characteristics across different scenarios are often not well understood. Moreover, the selected features typically capture only a fraction of the original channel information, potentially overlooking subtle and hidden channel attributes. This limitation becomes particularly evident in complex tasks, such as distinguishing nuanced channel scenarios where statistical characteristics exhibit similarities. As a result, existing statistical characteristics-based methods struggle to identify these scenarios.

To address these challenges, we consider the channel scenario identification relying on the channel estimation acquired by the communication transceiver. In specific, the scenario identification problem is formulated as the maximum a posteriori (MAP) estimation given the estimated channel. It is reformulated as the maximum likelihood estimation (MLE) via the Bayes theorem, where the likelihood function is implicit. Thanks to the powerful data-driven artificial intelligence method, the likelihood function is obtained via the conditional diffusion model \cite{SDM}. To the best of our knowledge, this work is the first investigation of exploiting generative diffusion models for channel scenario identification. Specifically, the reverse process of the conditional diffusion model is employed to model the underlying distributions of channels from different scenarios, where the latent features can be exploited via model training. 
Simulations demonstrate that the proposed method achieves striking results compared to conventional approaches, including BPNNs \cite{BPNN2021}, CNNs \cite{CNN}, and random forest \cite{RF}-based methods.


\section{Preliminaries}
In this section, we first formulate the problem of channel scenario identification, and then provide an overview of the conditional diffusion model.
\subsection{Channel Scenario Identification}
In this paper, we consider an identification task involving $C$ distinct channel scenarios, which can be formulated as a MAP estimation problem
\begin{align}
\label{MAP}
c^*\left(\mb{h}\right) = \argmax_{c \in \mathcal{C}} \text{ } p\left(c|\mb{h}\right).
\end{align}
Here, $\mathcal{C}$ is the set of considered scenarios, and $p\left(c|\mb{h}\right)$ is the posterior probability. $\mb{h}$ is the channel frequency impulse estimated by receivers and utilized for scenario identification in this paper. Moreover, the posterior probability $p\left(c|\mb{h}\right)$ can be considered as the probability that $\mb{h}$ belongs to a specific channel scenario $c$. However, due to the inherent complexity of the relationship between channels and their corresponding geographical environments, deriving a closed-form mathematical expression for $p\left(c|\mb{h}\right)$ is challenging. 
Therefore, based on the Bayes theorem, the posterior probability can be rewritten as
\begin{align}
\label{Bayes}
p\left(c|\mb{h}\right) & = \frac{p\left(c\right)p\left(\mb{h}|c\right)}{p\left(\mb{h}\right)}, \\
& \propto p\left(\mb{h}|c\right), \label{propto}
\end{align}
where $p\left(\mb{h}|c\right)$ is the likelihood function. The proportionality in (\ref{propto}) holds under the given $\mb{h}$ and a uniform prior $p\left(c\right) = \frac{1}{C}$, which is a natural assumption.
Furthermore, the MAP estimation in (\ref{MAP}) can be transformed into a MLE problem
\begin{align}
\label{MLE}
c^*\left(\mb{h}\right) = \argmax_{c \in \mathcal{C}} \text{ } p\left(\mb{h}|c\right).
\end{align}
Since the likelihood function $p\left(\mb{h}|c\right)$ remains intractable, we resort to the conditional diffusion model to approximate $p\left(\mb{h}|c\right)$ via data-driven training in this paper. To facilitate model training, we assume that there are $N_c$ channel samples collected from scenario $c$. The training set of scenario $c$'s channels denotes as $\mathcal{H}_c = \left\lbrace \mb{h}^{c,1}, \cdots, \mb{h}^{c,N_c} \right\rbrace $.

\subsection{The Conditional Diffusion Model}\label{2B}
The conditional diffusion model still adheres the structure of a Markov chain with $T$ time steps. In the fixed forward process, the corresponding latent samples $\mb{x}_t, t=1,\cdots, T$ can be obtained by progressively adding standard Gaussian noises into the original sample $\mb{x}_0$. Conversely, the reverse process aims to gradually remove the added noise to reconstruct the original sample. However, unlike standard diffusion models, the conditional diffusion model incorporates a conditioning mechanism in the reverse process, guiding noise removal by learning target distributions conditioned on $c$. 

Formally, the forward process generates latent sample $\mb{x}_t$ as described in
\begin{align}
\label{forward1}
\mb{x}_t & = \sqrt{\alpha _t} \mb{x}_{t-1} + \sqrt{1-\alpha_t}\epsilon_{t-1}, \\
\label{forward0}
& = \sqrt{\bar{\alpha}_t}\mb{x}_0 + \sqrt{1-\bar{\alpha}_t} \epsilon_t,
\end{align}
where $\epsilon_i \sim \mathcal{N}\left(\mb{0},\mb{I} \right),i=1,\cdots,t$ and $\bar{\alpha}_t = \prod _{i=1}^t {\alpha}_i$. From (\ref{forward0}), it can be seen that $\mb{x}_{t}$ adheres to a Gaussian distribution $ \mathcal{N}\left(\sqrt{\bar{\alpha}_t}\mb{x}_0, (1-\bar{\alpha}_t) \mb{I} \right)$.

Furthermore, the reverse process in conditional diffusion models seeks to sample from a learned target distribution $g_{\theta}$ by progressively removing white Gaussian noise. The process of recovering $\mb{x}_0$ conditioned on $c$ is defined as
\begin{align}
\label{inverse}
g_{\theta} \left( \mb{x}_0 | c \right) = \int_{\mb{x}_{1:T}}{ p\left(\mb{x}_T\right) {\prod_{t=1}^{T} g_{\theta} \left( \mb{x}_{t-1} | \mb{x}_{t},c \right) d\mb{x}_{1:T}} },
\end{align}
where $p\left(\mb{x}_T\right)$ typically follows a white Gaussian distribution $\mathcal{N}\left(\mb{0},\mb{I} \right)$. However, $g_{\theta} \left( \mb{x}_0 | c \right)$ is intractable to be trained directly due to the multi-fold integrations. Instead, following \cite{SDM}, the conditional diffusion model is trained by optimizing the variational lower bound of the log-likelihood, as
\begin{align}
\label{Loss}
\log g_\theta \left( \mb{x}_{0} |c \right) 
\ge
\sum\limits_{t=1}^{T} \mc{D}_{KL} \left( q \left( \mb{x}_{t-1} | \mb{x}_{t}, \mb{x}_0 \right) \| g_{\theta} \left( \mb{x}_{t-1} | \mb{x}_{t},c \right)   \right).
\end{align}
Here, $q\left( \mb{x}_{t-1} | \mb{x}_{t}, \mb{x}_0 \right)$ and $g_{\theta} \left( \mb{x}_{t-1} | \mb{x}_{t},c \right)$ represent the actual reverse process of (\ref{forward1}) given $\mb{x}_0$ and the approximate reverse process fitted by a model, respectively. $\mc{D}_{KL}$ denotes the Kullback-Leibler (KL) divergence, which quantifies the difference between $q\left( \mb{x}_{t-1} | \mb{x}_{t}, \mb{x}_0 \right)$ and $g_{\theta} \left( \mb{x}_{t-1} | \mb{x}_{t},c \right)$. 
Furthermore, authors of \cite{CDM} reparameterized $g_{\theta} \left( \mb{x}_{t-1} | \mb{x}_{t},c \right)$ by a Gaussian $\mathcal{N}\left( \mb{\mu}_{\theta} \left(\mb{x}_t, c \right),\mb{\beta}_{\theta} \left(t\right)\mb{I} \right)$, where $\mb{\mu}_{\theta} \left(\mb{x}_t, c \right)$ is the learned mean and $\mb{\beta}_{\theta} \left(t\right)\mb{I}$ is the constant covariance matrix that depends on time.
Since both $q$ and $g_\theta$ are Gaussian distributions, $\mc{D}_{KL}$ can be computed by a Rao-Blackwellized method with closed form expressions. Similar to processes in \cite{CDM}, $\mc{D}_{KL} \left( q \left( \mb{x}_{t-1} | \mb{x}_{t}, \mb{x}_0 \right) \| g_{\theta} \left( \mb{x}_{t-1} | \mb{x}_{t},c \right)   \right)$ is simplified as 
\begin{align}
\label{Ltrain}
- \mbb{E}_{t,\epsilon_t} \left[\left\| \mb{\epsilon}_{t} - \mb{\epsilon}_{\theta} \left(\mb{x}_t, c \right) \right\|_2^2 \right],
\end{align}
where $\mb{\epsilon}_{t}$ represents the actual sample noise at time step $t$, while $\mb{\epsilon}_{\theta} \left(\mb{x}_t, c \right)$ is the noise predicted by the model $ \mb{\epsilon}_{\theta} $. $\left\| \cdot \right\|_2^2 $ denotes the square of its 2-norm. In this paper, $\mb{x}_0$ corresponds to the channel frequency impulse $\mb{h}$, while $c$ represents the channel scenario label.

\section{Proposed Conditional Diffusion Model-Based Scenario Identification Algorithm} \label{sec3}
\begin{figure}[t]
\centerline{
\includegraphics[width=3.15in]{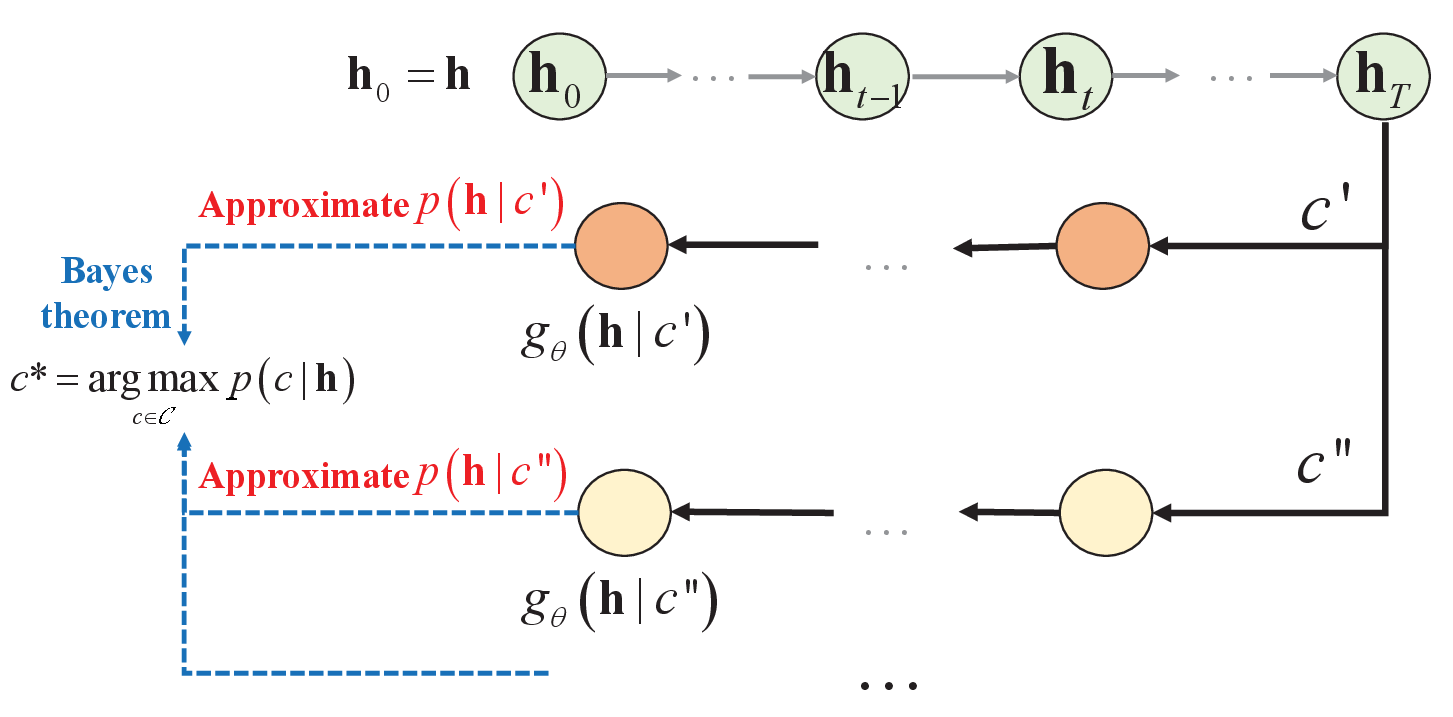}}
\caption{The pipeline of the proposed scenario identification algorithm.}
\label{Mov}
\end{figure}
In this section, we will discuss the derivation of the proposed scenario identification algorithm. The pipeline of the proposed scenario identification algorithm is shown in Fig. \ref{Mov}. First, the intractable likelihood function $p\left(\mb{h}|c\right)$ in MLE is approximated as a Gaussian distribution. This is then further modeled as the reverse generation process of the conditional diffusion model, denoted by $g_\theta\left(\mb{h}|c\right)$. Therefore, (\ref{MLE}) can be approximated as the following problem
\begin{align}
\label{MLEg}
c^*\left(\mb{h}\right) = \argmax_{c \in \mathcal{C}} \text{ } g_\theta \left(\mb{h}|c\right).
\end{align}

Moreover, as described in subsection \ref{2B}, training $g_\theta\left(\mb{h}|c\right)$ is converted into training the noise prediction model $\mb{\epsilon}_{\theta}$. Based on (\ref{Loss}) and (\ref{Ltrain}), $g_\theta \left(\mb{h}|c\right)$ can be calculated by the variational lower bound of the log-likelihood, and is represented by a uniform  distribution
\begin{align}
g_\theta \left(\mb{h}|c\right) = \frac{\exp \left( - \mbb{E}_{t,\epsilon_t} \left[\left\| \mb{\epsilon}_{\theta} \left(\mb{h}_t, c \right) - \mb{\epsilon}_{t} \right\|_2^2 \right] \right) } {\sum_{c'=1}^{C} \exp \left\lbrace- \mbb{E}_{t,\epsilon_t} \left[\left\| \mb{\epsilon}_{\theta} \left(\mb{h}_t, c' \right) - \mb{\epsilon}_{t} \right\|_2^2 \right] \right\rbrace }.
\end{align}
To obtain an unbiased estimation of $g_\theta \left(\mb{h}|c\right)$, we employ Monte Carlo techniques to compute the expectation over $M$ samples. Given $t_m \sim \left\lbrace 1,T \right\rbrace $ and $\epsilon_{t_m} \sim \mathcal{N}\left(\mb{0},\mb{I} \right)$, the expectation is computed as
\begin{align}
\label{E}
\mbb{E}_{t,\epsilon_t} \left[\left\| \mb{\epsilon}_{\theta} \left(\mb{h}_t, c \right) - \mb{\epsilon}_{t} \right\|_2^2 \right] = \frac{1}{M} \sum_{m=1}^{M} \left\| \mb{\epsilon}_{\theta} \left(\mb{h}_{t_m}, c \right) - \mb{\epsilon}_{t_m} \right\|_2^2, 
\end{align}
where
\begin{align}
\label{tm}
\mb{h}_{t_m} = \sqrt{\bar{\alpha}_{t_m}}\mb{h} + \sqrt{1-\bar{\alpha}_{t_m}} \epsilon_{t_m}.
\end{align}

\begin{figure}[t]
\centerline{
\includegraphics[width=2.85in]{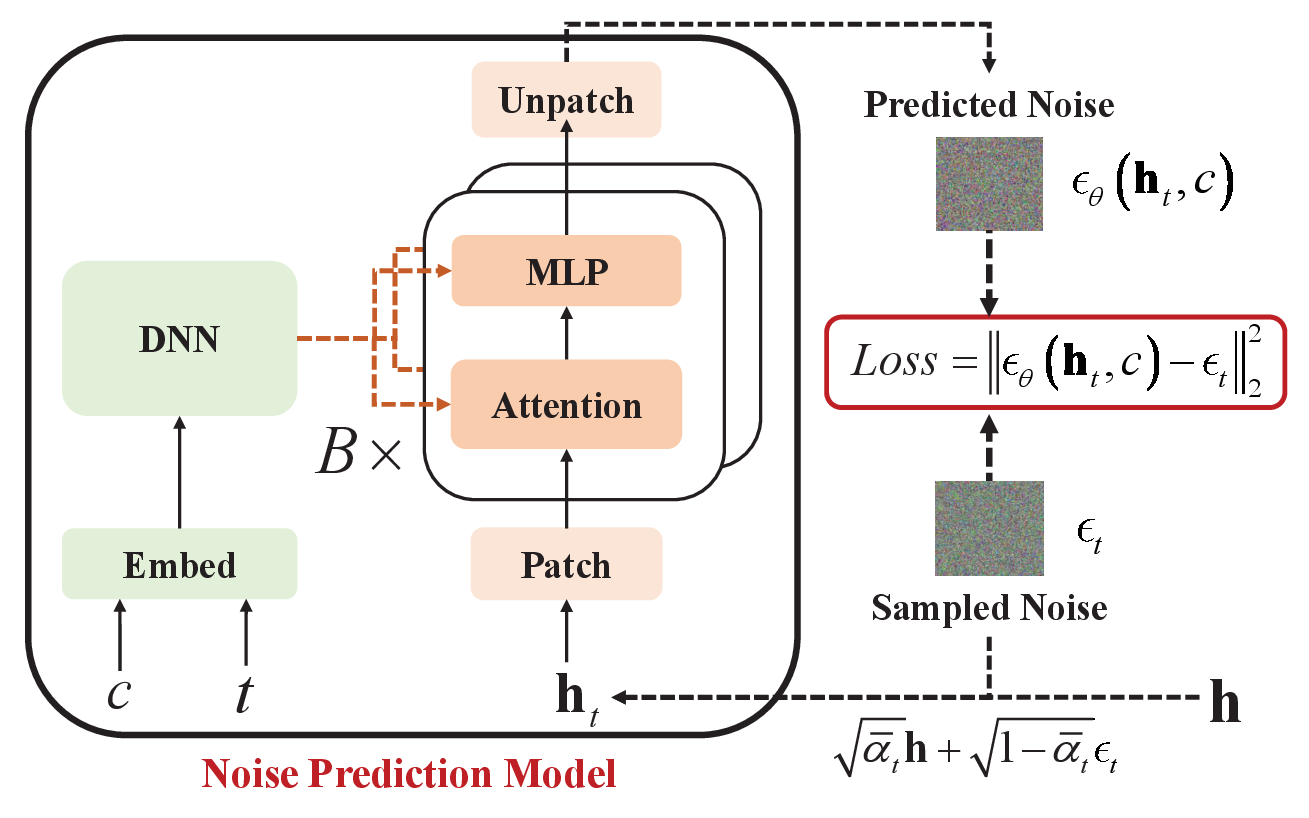}}
\caption{The noise prediction model.}
\label{NoiseModel}
\end{figure}

To train the noise prediction model $\epsilon_\theta$, we design a transformer-based network, as illustrated in Fig. \ref{NoiseModel}. The input to the noise prediction model is the noisy channel $\mb{h}_t$, generated through the forward process using the original channel $\mb{h}$ and sampled ground truth noise $\epsilon_t$. The designed noise prediction model consists of three main components, i.e., a patch layer, a transformer layer, and an unpatch layer.
In the patch layer, the multi-dimension input $\mb{h}_t$ is transformed into token sequences. These sequences are then processed by the transformer layer to extract hidden features. The transformer layer includes $B$ blocks, each containing a cross-attention network and a multi-layer perceptron (MLP). Notably, partial related parameters of the transformer layer, including the Key and Value of the cross-attention network, along with the weights of the MLP's parameters, are conditioned on the scenario label $c$ and the time step $t$. Specifically, as shown in green parts of Fig. \ref*{NoiseModel}, $c$ and $t$ are first embedded as conditional features and subsequently input into a deep neural network (DNN) to generate partial related parameters of the transformer layer.
Based on this structure, the cross-attention network dynamically captures global dependencies of channels guided by $c$ and $t$. Furthermore, by extracting nonlinear hidden features, the MLP dynamically enhances local feature representations of channels under the conditions $c$ and $t$. Consequently, different global and local features will be mined conditioned on different scenario label $c$.
Finally, the unpatch layer maps the transformer network's output to the predicted noise. The loss function of the noise prediction model is defined as
\begin{align}
\label{Lossfunc}
Loss = \left\| \mb{\epsilon}_{t} - \mb{\epsilon}_{\theta} \left(  \mb{h}_t, c \right) \right\|_2^2.
\end{align}

The training procedure for the proposed scenario identification algorithm is summarized in Algorithm \ref{algtrain}. Step 1 defines the required inputs for training. Steps 4 to 7 generate a noisy channel $\mb{h}_t$ with known scenario condition $c$ and ground truth noise $\epsilon_t$. In step 8, $\mb{\epsilon}_{\theta} \left(\mb{h}_t, c \right)$ is predicted by the designed network. The loss function is computed in step 9, followed by parameter updates using gradient descent in step 10. As the number of samples increases, the loss function converges, and the noise prediction model $\epsilon_\theta$ is well-trained.

\begin{algorithm}[t]
	\caption{Training Procedures of Proposed Scenario Identification Algorithm}
	\label{algtrain}
	\begin{algorithmic}[1]
		\State  \textbf{Input:}	Scenarios set $\mathcal{C}$, training set of scenario $c$'s channels $\mathcal{H}_c = \left\lbrace \mb{h}^{c,1}, \cdots, \mb{h}^{c,N_c} \right\rbrace $, and fixed diffusion coefficient $\bar{\alpha}_{t}, t=1, \cdots, T$.
        \State  \textbf{Output:} The noise prediction model $\epsilon_\theta$.
		\Repeat
		\State Randomly sample $c$ from $\mathcal{C}$.
		\State Randomly sample $\mb{h}$ as $\mb{h}_0$ from $\mathcal{H}_c$.
		\State Randomly sample $t$ from $\left\lbrace 1,2,\cdots,T \right\rbrace $.
		\State Randomly sample $\epsilon_{t}$ from $\mathcal{N}\left(\mb{0},\mb{I} \right)$, and calculate $\mb{h}_t = \sqrt{\bar{\alpha}_{t}}\mb{h} + \sqrt{1-\bar{\alpha}_{t}} \epsilon_t$.
        \State Input $c$, $t$, $\mb{h}_t$ into the noise prediction model to predict $\mb{\epsilon}_{\theta} \left(\mb{h}_t, c \right)$.
		\State Calculate the loss via (\ref{Lossfunc}), and then update the parameters of the noise prediction model using the gradient descent step $ \nabla_{\theta}\left( \left\| \mb{\epsilon}_{\theta} \left(\mb{h}_t, c \right) - \mb{\epsilon}_{t} \right\|_2^2 \right) $.
		\Until The training stop condition is met.
	\end{algorithmic}
\end{algorithm}

Then, we outline the testing procedures in Algorithm \ref{algtest}. To facilitate the computation of the expectation over $M$ observations, we initialize variables $\eta_c = 0, c=1,\cdots,C$ in step 3. For the $m$-th observation, the time step $t_m$ and ground truth noise $\epsilon_{t_m}$ are sampled in steps 5 and 6. Then, the test channel $\dot{\mb{h}} $ is utilized to construct the noisy channel $\dot{\mb{h}}_{t_m}$  in step 7. Steps 8 to 11 compute the error between the predicted noise $\mb{\epsilon}_{\theta} \left(\dot{\mb{h}}_{t_m}, c \right)$ and the ground truth noise $\epsilon_{t_m}$ for each scenario $c$, updating $\eta_c$ accordingly. In these steps, the well-trained noise prediction model captures diverse local and global features across different scenarios, aiding in predicting different noise. However, the most accurate and comprehensive features are learned only under the correct scenario, resulting in predicting more precise noise. In steps 13 to 15, we calculate the generation distribution of each scenario. Finally, the scenario of $\dot{\mb{h}} $ is identified in step 16.

\begin{algorithm}[t]
	\caption{Testing Procedures of Proposed Scenario Identification Algorithm}
	\label{algtest}
	\begin{algorithmic}[1]
		\State  \textbf{Input:}	Scenarios set $\mathcal{C}$, testing channel $\dot{\mb{h}} $, fixed diffusion coefficient $\bar{\alpha}_{t}, t=1, \cdots, T$, and the well-trained noise prediction model $\epsilon_\theta$.
        \State  \textbf{Output:} Identified scenario condition $c^*\left(\dot{\mb{h}}\right)$.
        \State  Initialize $\eta_c = 0, c=1,\cdots,C$.
        \For{$m = 1:M$}
      	\State Sample $t_m$ from $\left\lbrace 1,2,\cdots,T \right\rbrace $.
      	\State Sample $\epsilon_{t_m}$ from $\mathcal{N}\left(\mb{0},\mb{I} \right)$.
      	\State Calculate $\dot{\mb{h}}_{t_m}$ based on the input $ \dot{\mb{h}}_0 = \dot{\mb{h}}$ by (\ref{tm}).
		\For{$c = 1:C$}
		\State Predict the noise $\mb{\epsilon}_{\theta} \left(\dot{\mb{h}}_{t_m}, c \right)$.
		\State $\eta_c = \eta_c + \left\| \mb{\epsilon}_{\theta} \left(\dot{\mb{h}}_{t_m}, c \right) - \mb{\epsilon}_{t} \right\|_2^2 / M $.
        \EndFor
        \EndFor
        \For{$c = 1:C$}
		\State $g_\theta\left(c|\dot{\mb{h}}\right) = \frac{  \exp \left( - \eta_c \right) }{ \sum_{c'=1}^{C} \exp \left( -\eta_{c'} \right) } $.
        \EndFor
		\State The scenario condition $c^*\left(\dot{\mb{h}}\right)$ is identified by (\ref{MLEg}).
	\end{algorithmic}
\end{algorithm}

\section{Numerical Results}
To evaluate the performance of the proposed scenario identification method, we analyze a concrete example using measured channels provided by \cite{Dataset}. There are three offices of identical proportions, yet their placement of obstacles differs, thus serving as three channel scenarios. In these scenarios, there is a transmitter in the same fixed position. Simultaneously, 800 positions are sampled to deploy receivers to estimate channels from the transmitter.
The measurement hardware includes two NI Universal Software Radio Peripherals (USRPs) 2974, acting as the transmitter and receiver, respectively. The transmission operates at a carrier frequency of 3.75 GHz with a bandwidth of 100 MHz, retaining 400 frequency points after excluding guard bands. The transmission power is set to 10 dBm. The measured channels are divided into training and test sets, and the sampling rate is defined as the ratio of training set samples to the total dataset size. The key hyperparameters of the conditional diffusion model, such as the number of time steps $T$ and the number of transformer blocks $B$, are summarized in Table~\ref{paset}.

\begin{table}[t]
\caption{Parameter setting of the conditional diffusion model}
\begin{center}
\begin{tabular}{|c|c|c|c|c|c|}
\hline
\textbf{Parameter} & \textbf{$T$} & \textbf{$B$} & Pitch size & Hidden size & Depth  \\
\hline
\textbf{Value} & 30 & 4 & 2 & 64 & 20 \\
\hline
\end{tabular}
\label{paset}
\end{center}
\end{table}

To validate the performance of the proposed scenario identification method, we compare it against three benchmarks, including the random forest algorithm from \cite{RF}, CNN from \cite{IndoorUWBCNN}, and the BPNN from \cite{BPNN2021}. Notably, in \cite{BPNN2021, IndoorUWBCNN, RF}, identification networks rely on selected statistical features extracted from channels, whereas our proposed method directly utilizes the original channels as input. To ensure a fair comparison, the input for both the proposed method and three baselines consists of concatenated data, comprising the original channels alongside four statistical features, i.e., path loss, delay extension, number of paths, and Rice K-factor.

\begin{figure}[t]
	\centering
	\includegraphics[width=2.8in]{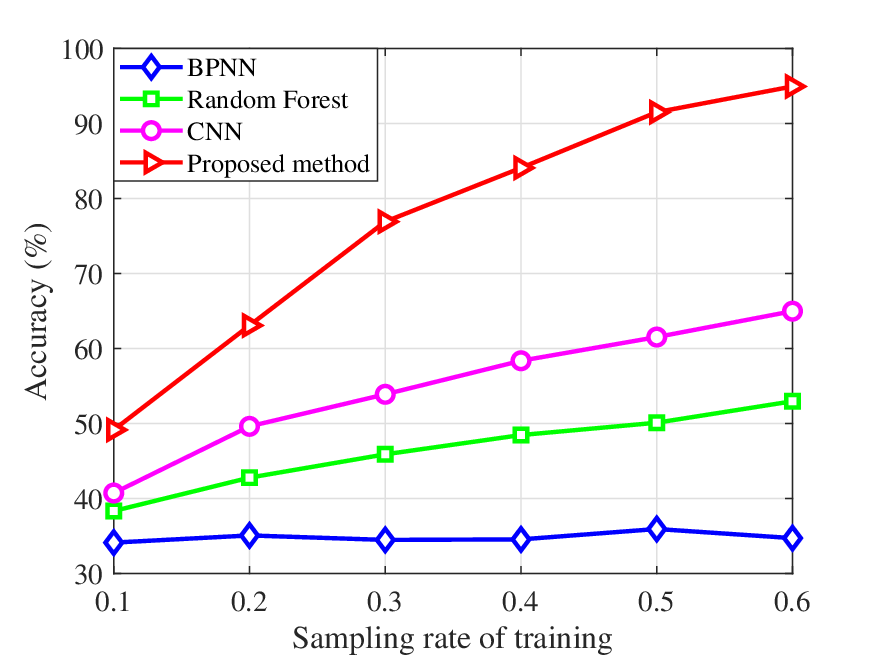}
	\caption{Identification performance with varying training sampling rates.}
	\label{Sam}
\end{figure}

Fig. \ref{Sam} presents the identification accuracy of the four scenario identification methods under varying sampling rates. The vertical axis represents the identification accuracy on the test set. As observed, the proposed method outperforms three benchmarks. Furthermore, the performance of the proposed method improves significantly as the sampling rate increases. While the CNN-based and random forest-based methods show improved accuracy with increasing sampling rates, the BPNN-based method struggles to distinguish between the three similar indoor scenarios. Because these benchmarks classify scenarios based on decision boundaries derived from statistical channel features, potentially overlooking additional critical characteristics. 
In contrast, the transformer network employed by the conditional diffusion model effectively captures hidden channel features in different noise spaces. By leveraging the difference of these hidden features, the proposed method achieves higher precision in scenario identification. 

\begin{figure}[t]
	\centering
	\includegraphics[width=2.8in]{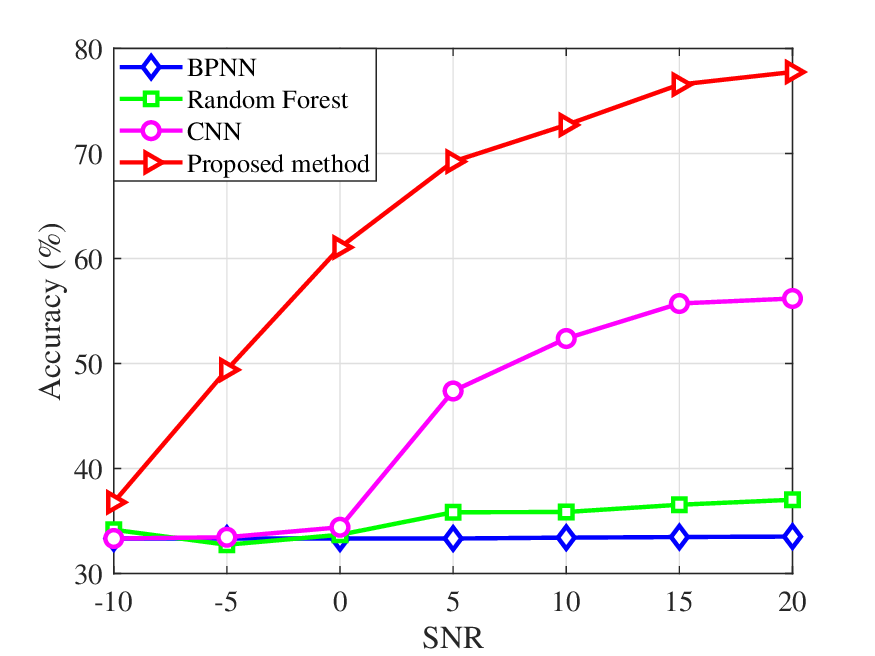}
	\caption{Identification performance with varying SNRs.}
	\label{SNR}
\end{figure}

Finally, at the sampling rate of 0.3, we compare the performance of the four scenario identification methods under varying signal-to-noise ratio (SNR) conditions. To generate test data for different SNRs, noise of varying power is superimposed on transmission signals during transmission, and channels are estimated using the LS method. As shown in Fig.~\ref{SNR}, the identification accuracy of the proposed method, as well as the CNN-based and BPNN-based methods, improves as the SNR increases. However, the proposed method consistently outperforms three benchmarks across the entire SNR range. Notably, at high SNR, the accuracy advantage of the proposed method becomes even more pronounced. These results demonstrate the superior robustness and effectiveness of the proposed method in scenario identification.

\section{Conclusion}
In this article, we proposed a high-precision channel scenario identification method implemented by a conditional generative diffusion model. The task of channel scenario identification was reformulated as a maximum likelihood estimation problem. Further, the maximum likelihood function was estimated by the conditional generative diffusion model. Unlike existing methods that relied on compressed statistical channel characteristics for scenario identification, our approach leveraged the powerful generative capabilities of the diffusion model to learn the underlying distributions of original channels from different scenarios.
By directly modeling the original channels, the inverse process effectively uncovered differentiated hidden features of channels specific to each scenario. These detailed features enable highly accurate and robust channel scenario identification.
Building on this work, we can explore scenario identification-assisted transceiver design to enhance the transmission performance of communication networks in the future.

\section*{Acknowledgment} 
This work was supported in part by the National Key Research and Development Program of China under Grant 2022YFB2902003; in part by the National Science Foundation of China under Grant 62471039; in part by the Guangdong Basic and Applied Basic Research Foundation under Grant 2023A1515012189; in part by the Guangdong Basic and Applied Basic Research Foundation under Grant 2022A1515110602; in part by the Guangdong Introducing Innovative and Entrepreneurial Teams of “The Pearl River Talent Recruitment Program” under Grant 2021ZT09X044; in part by the Guangdong Introducing Outstanding Young Scholars of “The Pearl River Talent Recruitment Program” under Grant 2021QN02X546.

\end{document}